# Aspect-based Opinion Extraction from Customer Reviews

Amani K Samha, Yuefeng Li and Jinglan Zhang

School of Electrical Engineering and Computer Science, Queensland University of Technology, Brisbane, Australia
amani.samha@student.qut.edu.au
{y2.li, jinglan.zhang}@qut.edu.au

## ABSTRACT

*Text is the main method of communicating information in the digital age. Messages, blogs, news articles, reviews, and opinionated information abounds on the Internet. People commonly purchase products online and post their opinions about purchased items. This feedback is displayed publicly to assist others with their purchasing decisions, creating the need for a mechanism with which to extract and summarize useful information for enhancing the decision-making process. Our contribution is to improve the accuracy of extraction by combining different techniques from three major areas, named Data Mining, Natural Language Processing techniques and Ontologies. The proposed framework sequentially mines product's aspects and users' opinions, groups representative aspects by similarity, and generates an output summary. This paper focuses on the task of extracting product aspects and users' opinions by extracting all possible aspects and opinions from reviews using natural language, ontology, and frequent "tag" sets. The proposed framework, when compared with an existing baseline model, yielded promising results.*

## 1. INTRODUCTION

The Internet contains vast amounts of textual information on people's expressed opinions, making the Internet an excellent source from which to gather data about a specific object within a specific domain. The ubiquity of customers' posted feedback has triggered the urgent need for systems that can automatically summarize documents. Searches for information about items available for purchase return enormous quantities of information, making it difficult to find useful data easily. Useful online information needs to be presented in a summarized form that includes the relevant data in easy-to-read and easy-to-understand format.

Reviews, forums, discussion groups, and blogs available on the Internet contain opinions and opinionated information. If extracted and summarized, those opinions could provide useful data for decision makers. The process of summarizing opinions relies primarily on identifying and extracting vital opinionated information from text. Efficiency of the process and quality of the resulting summary depends on the extraction of key information and exclusion of superfluous details. Both individuals and businesses seek opinion summaries to enhance their decision-making processes.

Feedback about purchased items can be objective and factual or subjective and opinionated. One customer's opinions may not fully represent the opinions of all customers, underscoring the importance of collecting and analysing opinions from many different opinion holders to evaluate the object under study. The need to understand customers' subjective feedback has made opinion extraction and summarization a hot subject in recent years. In opinion summarization, opinions are extracted, analysed, summarised, and then presented along with the corresponding opinionated information.

Researchers have studied various types of extraction and summarization, as well as methods to create and evaluate the final summary. This paper reviews recent work and covers some techniques on extracting and summarizing opinions. The primary focus is analysing customers' opinionated reviews, extracting opinionated aspects by applying the proposed framework to present extracted knowledge as "aspect-based opinion summary". The aim of this study is to achieve this goal by improving the accuracy of the aspect-based opinion summarization model to improve the quality of opinion summarization from customers' reviews. This paper documents development of a new technique to extract product aspects along with consumers' opinions about those products and aspects with the use of data mining techniques, natural language processing and ontologies. We begin with a discussion of some related work, followed by an explanation of the proposed framework, then the proposed extraction techniques, followed by experiment and evaluation, and finally conclusion with some recommendations for future work.

## 2. RELATED WORK

Opinion summarization from online customer reviews mainly consists of three tasks. First, aspects must be extracted. Then, associated opinion must be identified and oriented. Finally, sentence lists must be produced to form the final summary. The effectiveness of the final summary relies on aspect identification and extraction. Opinion is a perspective or a judgment formed about something; opinion is not necessarily based on fact or existing knowledge [1]. Conducting sentiment analysis is problematic [2, 3] because opinion is a quintuple of entity, aspect, orientation, opinion holder, and time [4]. The entity is the item being studied (e.g., a product). The aspect can be feature, component, or function of the entity. While, orientation is the opinion provided about the entity and/or the aspect that was provided by the opinion holder at a specific time.

Summary is another concept of interest related to opinions; as explained in [5], a summary is "text that is produced from one or more texts, that conveys important information in the original text [6], which is no longer than half of the original text/s and usually significantly less than that. The Oxford Dictionary [1], defines summary as "a brief statement or account of the main points of something", and defines sentiment as "an exaggerated and self-indulgent feelings of tenderness, sadness, or nostalgia" [1].

Four broad categories of feedback for entities represent the types of words most frequently used: components, functions, features, and opinions [7, 8]. Entities for "camera" are demonstrated in Table 1. Some entities do not fit into any of the four established categories, so a fifth category, "other," is used to capture these terms.

Table 1. Entity Categories

| Entity | Description |
| --- | --- |
| Components | Physical aspects, including the camera itself, LCD, viewfinder, battery |
| Functions | Capabilities, including movie playback, zoom, and autofocus |
| Features | Properties of components or functions, such as colour, speed, size, weight, and clarity |
| Opinions | Ideas and thought expressed by reviewers on product, features, components, or functions |
| Other | Other possible entities defined by the domain |

To date, most methods have focused on extracting product aspect/features from online customer feedback and then summarizing the results, which is the first step to produce an opinionated aspect-based summary of the product under study. Hu and Liu [2, 3] presented a novel technique that performs extraction and summarization of customer reviews by using association rules based on an a priori algorithm. The system that Hu and Liu designed, extracted frequently used words representing aspects/features. In 2005, [9] proposed a modified version of the original system based on language pattern mining that identified explicit and implicit product aspect/features from positive and negative reviews.

Carenini et al. in [10] sought to improve the aspect extraction of prior designs using output from Hu and Liu's [2] model as input to their system to capture knowledge from customer reviews. The model worked by mapping the input to the user-defined taxonomy of the aspect hierarchy to eliminate redundancy and provide conceptual organization. Yi and Niblack in [11] developed a set of aspect extraction heuristics and selection algorithms to extract aspect from reviews. This model worked by extracting noun phrases, then selecting feature terms using likeness scores [12]. Popescu and Etzioni in [13] made more improvements to Hu and Liu's work [2, 3] by developing an unsupervised information system that extracted product aspects and opinions by mining reviews and removing frequently appearing nouns that are not aspects. The result of this improved system was increased precision but low recall compared to previous work.

Wu et al. in [14] proposed a novel approach to identify noun and verb phrases as aspect/features and opinion expressions, and then find the relationships between them. The method worked by extending traditional dependency parsing to the phrase level, which worked well in mining. Qiu et al. in [15] took a different approach by focusing on extraction of nouns and noun phrases, and then finding relationships between opinion words and target expressions based on dependency parsing. Both of these methods achieved normal recall performance and low precision but failed to extract infrequently cited aspects.

In [16], Qi and Chen proposed a discriminative model by using linear-chain conditional random files to mine opinions. Results of this model yielded improvements in recall and precision compared to other methods proposed by Turney [17] and Jin et al. [18]. Huang et al. [19]proposed aspect/feature extraction as a sequence label by implementing the discriminative learning model. This approach performed well, achieving an increase in both recall and precision.

## 3. PROPOSED FRAMEWORK

The proposed framework was designed to summarize customer reviews and produce "aspect - based opinion summary". To produce a representative summary, some essential information must be extracted. The framework is divided into four major tasks to use text files containing customer reviews as input and then perform the four tasks to produce the final output summary.

The first task is to mine entities (aspects and opinions) of the product under study and identify the associated opinion orientation of each aspect. The second task is to group aspects based on similarities. The third task is to select the most popular aspect sentences. The fourth task is to generate an opinionated summary that is based on product aspects. The architecture of the proposed framework is shown in Figure 1.

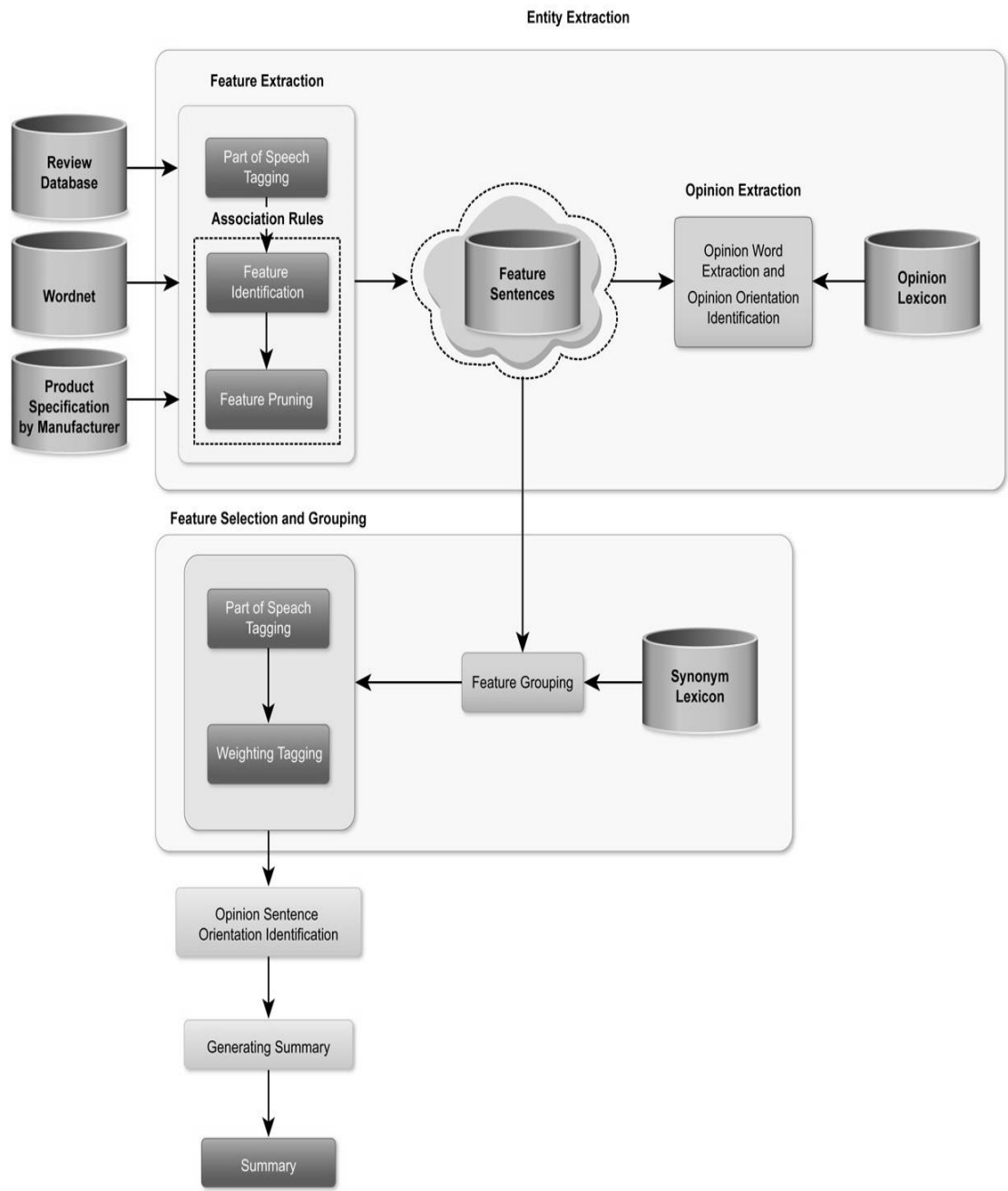

Figure 1. Proposed Framework

Although we touch on the four tasks, the focus of this paper is the proposed technique by which all possible aspects are extracted from customer reviews.

### 3.1. Entity Extraction

The first task of the proposed framework is "entity extraction". According to [7], entities include aspects/features, components, parts, functions, and opinions of the object being studied. For our work, entity extraction is handled as two extractions: product aspects extraction and opinions extraction. Furthermore, the extraction of aspects is decomposed into two-step process.

## 3.2. Aspect Grouping

Once entities have been extracted, they are grouped by based on synonyms. People may express their opinions about the same aspect using different words and/or phrases. To produce a useful summary, those different words about the same aspect must be grouped. Those words and/or phrases are domain synonyms—they share the same meaning and so must group them under the same aspect group. In a mobile phone domain, for instance, *"capacity"* and *"memory"* are two different expressions referring to the same aspect.

In this paper, aspect grouping is critical due to the numerous possible synonyms. The level of sufficiency is low for two reasons. First, although words may refer to the same aspects, some dictionaries do not consider words to be synonyms. Second, many synonyms are domain synonyms; they are likely to refer to the same aspect in one domain but not in another [20].

We aim to achieve aspect grouping using natural language possessing techniques, shared words and lexicon similarity. Some aspects may share words e.g., (*"battery,"* *"battery life,"* *"battery usage,"* and *"battery power"*), all of which refer to the same aspect—*"battery"* [20]. Moreover, using lexicon similarity, we will match the extracted aspects to WordNet dictionary to obtain synonyms[21, 22].

## 3.3. Aspect Selection

After aspects have been grouped, the most representative aspect sentences must be selected to form the final opinionated summary. This step can be accomplished by analysing the strength of each opinionated sentence and then select sentences with the highest weight. The strength of all "adjectives, adverbs and verbs ", within the sentence, will determine the total weight of that sentence. Sentence importance is one of the most critical determinations of this proposed framework.

In this paper, we calculate the weights for all "adjectives, adverbs and verbs "for each the sentence. The calculation is done by adding up all weights for each "adjectives, adverbs and verbs "within the sentence, as presented in Table 2. For example, *"earpiece is very comfortable"*, the sentence has an "adjective = *comfortable*" and "adverb = *very*", therefore, the earned weight for this sentence is "2".

The weights are calculated based on the a method to score a combination of tags (adjective, verb, adverbs) to give weight to each aspect sentence, as indicated in Table 2 for adjectives and adverbs and Table 3 for verbs based on the approach proposed by [23].

Table 2. Adjective and adverb weights

| Tags | Description | Weight |
| --- | --- | --- |
| JJ | Adjective | 1 |
| JJR | Comparative Adjective | 2 |
| JJS | Superlative Adjective | 3 |
| RB | Adverb | 1 |
| RBR | Comparative Adverb | 2 |
| RBS | Superlative Adverb | 3 |

On other hand, verbs are treated differently from adjectives and adverbs. We used the categories proposed by [23] to weigh verbs, some categories are shown in Table 3. If the sentence contains a verb from positive categories, then "+1" will be added to the weight and if the verb is from negative categories then "-1" will be subscribed form the total weight. Based on final sentence's weights, the selection can be easily made. We will select sentences with the highest weight to be candidatures for the final summary.

Table 3. Verb weights

| Verb category | Orientation | Verbs | Comments |
|---|---|---|---|
| Tell verbs | Positive | tell | Positively reinforce an opinion |
| Chitchat verbs | Positive | argue, chatter, gab | Positively reinforce opinion is being expressed |
| Advise verbs | Positive | advise, instruct | Positively reinforce an opinion |
|  | Negative | admonish, caution, warn | Negatively reinforce the degree of certainty about a given opinion |

### 3.4. Summary Generation

Summary generation is the final task of the process. It is based on the outcomes of the preceding tasks in which the extracted aspects and its corresponding opinion are selected and then weights are given to all sentences. The summary could be presented in various forms, such as diagram, text, or graph. Our expected output summary takes the form of pros and cons along with a horizontal histogram, where the pros indicate the set of positive product aspects/opinions and the cons represent the set of negative aspects/opinions. The horizontal histogram included as the percentage of positive opinions compared to negative opinions for all sentences. Figure 2 is an example, of an aspect-based summary of "*MP3 player*".

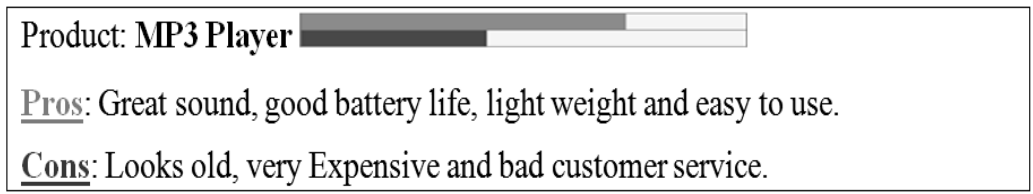

Figure 2. Aspect-based opinion summary

### 4. PROPOSED EXTRACTION TECHNIQUE

As illustrated in previous sections, system input is a list of customers' reviews of a specific product and the output is a summary of all reviews of that product. The initial tasks of this paper rely on part-of-speech (POS) tagging.

### 4.1. Part-of-Speech (POS) Tagging

To extract useful information such as aspects and opinions from reviews, the reviews must be parsed and parts of speech tagged accordingly. Part-of-speech (POS) tagging is the process of parsing each word of the sentence based on identifying linguistic tags. Table 4 shows a list of linguistic POS tags. To illustrate the use of POS tagging, we offer the example of a customer's review of an iPhone5s. The original sentence is, *"I love my new Iphone5s, it is the best Smartphone ever, and it has a great camera that captures the best photos."* The tagged sentence is *"I/PRP love/VBP my/PRP$ new/JJ IPhone/NN 5s/NNS, /, it/PRP is/VBZ the/DT best/JJS smartphone/NN ever/RB, /, it/PRP has/VBZ a/DT great/JJ camera/NN that/WDT captures/VBZ the/DT best/JJS photos/NNS /"* where every word is tagged using the categories shown in Table 4.

Table 4. Part-of-speech (POS) tagging

| Tag | Description | Tag | Description |
|---|---|---|---|
| JJ | Adjective | RBR | Comparative adverb |
| JJR | Comparative adjective | RBS | Superlative adverb |
| JJS | Superlative adjective | VB | Verb, base form |
| LS | List item marker | VBD | Verb, past tense |
| NN | Noun, singular or mass | VBG | Verb, gerund, or present participle |
| NNS /NNP | Noun, plural noun, singular | VBN | Verb, past participle |
| NNPS | Proper noun, plural | VBP | Verb, non-3rd-person singular/p |
| RB | Adverb | VBZ | Verb, 3rd-person singular present |

Earlier research [2, 3] demonstrated that product aspects tend to be nouns or/and noun phrases and opinions tend to be adjectives or/and adjective phrases. In [23], sentiment analysis research showed that some combination of tags contribute to aspects and opinion extraction. Unlike these previous studies, the current research made more use of the sentence parsing process by considering more parts of the sentence to be aspects or/and opinions.

The proposed framework is designed to determine what people like and dislike about a given product. Identifying the aspects of this product is the first task, followed by finding the corresponding opinions. Understanding natural language is not easy, so the extraction process is not easy as well. The major difficulty is to understand the implicit meaning of a specific sentence. For example, *"using Iphone5 is a piece of cake,"* the phrase *"piece of cake"* means it is easy to use. However, there is no explicit word to show that hidden meaning. To solve such issues, semantic understanding is needed.

In this paper, we use OpenNLP, part of the SharpNLP Project package[24], which is a collection of natural language processing (NLP) tools that are written in C# programming language. For semantic understanding, we used a linguistic parser tool included as part of SharpNLP, OpenNLP, which parses each sentence of the review and yields the tags of each word (noun, adjective, and so on). As Table 4 shows all the POS tags taken from the Penn Treebank project POS tags [25]. An additional tool, we used a WordNet database, SharpWordNet, to find synonyms in order to expand the aspect list. We use the produced output file from SharpWordNet to feed the proposed framework .

### 4.2. Product Aspects Extraction

Aspect extraction involves extracting aspects of the product being studied about which customers have expressed their opinions on. Aspects are usually nouns or/and noun phrases, for example, *"face recognition"*, *"zoom"*, and *"touch screen"* are aspects of the product *"camera"*. To extract aspects, we must analyse all review sentences to know which POS items presented as aspects and which presented as opinions.

In natural language, people tend to write almost similar sentence structure. From here, we choose to use frequent sets based on its success in analysing and understanding customer purchasing behaviour. Mining frequent sets plays a great role in data mining, it aims to find interesting patterns form large amount of data. Frequent sets were introduced by [26] to analyse customer behaviour and how customers tend to purchase sets of items together. The main motivation to search frequent "tag" sets, came from the need to analyse how people tend to express their feelings in natural language. In other words, how people tend to write opinionated reviews.

To achieve the maximum number of possible aspects, we first build a list of aspects obtained from the product specifications and expand the list by word synonyms. Product specifications are aspects of the product provided by the manufacturer, while synonyms are derived from the WordNet dictionary [21]. We apply POS tagging technique to 260 sentences, then we analysis the tags based on manual observations. In order to determine how people tend to write their opinionated reviews. Then we apply opinion lexicon to match opinion words to which tags it expressed. From there, we look up for aspects by engaging the list of aspects and its synonyms.

The output is frequent sets, which consisted of frequent tags that define the product aspects, the opinion words and the relationship between those two tags. For instance, the tag of aspect appears first, therefore, the sequent of tags looks like *[NN][VBZ][RB][JJ]* which correspond to the sentence *"software is absolutely terrible"* . Figure 3 and Figure 4 show tags that are more frequent, whereas Figure 5 shows how those tags are extracted.

- **[NN] [VBZ] [RB][JJ]** *e.g. "software is absolutely terrible"*
- **[NNS][VBP] [JJ]** *e.g. "pictures are razor-sharp"*
- **[NN][VBZ][RB][JJ]** *e.g. "earpiece is very comfortable"*
- **[NN] [VBZ] [JJ]** *e.g. "sound is wonderful"*
- **[NNS] [VBP] [RB]** *e.g. "transfers are fast"*
- **[VBZ][JJ]** *e.g. "looks nice"*

Figure 3. Frequent tags" Aspect appears first"

- **[JJ][NN] [IN] [NN]** *e.g. " superior piece of equipment"*
- **[JJ] [NN] [CC] [NN]** *e.g. "decent size and weight"*
- **[RB][JJ][TO][VB] [DT] [NN]** *e.g. "very confusing to start the program"*
- **[VBD] [NN]** *e.g. " improved interface"*
- **[JJ] [VBG]** *e.g. " great looking"*

Figure 4. Frequent tags" opinion appears first"

```
Algorithm AspectTagsExtraction ()
 //Input:    Sentences - List of sentences
           Dict - Feature Dictionary
           PSL - Positive Seed List
           NSL - Negative Seed List
 //Output:
           F1 - File Consisting of Possible features
           F2 - File Consisting: list of Feature & Opinion & sentence rows
2. for each sentence si ∈ Sentence do
3.    W = tokenize each word ∈ si  /*Tokenized sentence */
4.    T = tag each word ∈ si       /*Tagged sentence */
5.    for each Wi ∈ si do
6.       if Wi ∈ Dict then
7.          apply_TwoRuleTag(si, PSL, NSL, Dict, W, T, index);  //index of the current token in Wi
8.       else if Wi+1 ∈ Dict then
9.          apply_ThreeRuleTag(si, PSL, NSL, Dict, W, T, index);
10.      else if Wi+2 ∈ Dict then
11.         apply_FourRuleTag(si, PSL, NSL, W, T, index);
12.      end if
13.   end for
14. end for
```

Figure 5. Frequent tags extraction

### 4.3. Opinion Words Extraction

The second task of the extraction process is opinion extraction. This task involves extracting corresponding opinion words that customers used for every product aspects. Opinion words are usually adjectives that describe or express what customers think about product aspects. Usually, opinion words are located near aspects in the sentence. Some researches located opinion words as the closest adjective to the aspects [2, 3]. Nevertheless, we first locate the opinions words in the sentence and from there we determine the corresponding aspects by searching the sentence backwards firs for the closest aspect, if we did not find, then we search forwards.

In this paper, we use the opinion lexicon developed by Hu and Liu in [2, 3] to extract opinion words. It contains 6,800 positive and negative words in two different text files. If the word in our sentence matches the positive dictionary, the word is positive, and if a word matches the negative dictionary, then it is negative. Then, the weights for adjective are given based Table 2. Then we apply the frequent sets of tags to validate the relationship between the opinion word and the aspects. The extraction algorithm is shown in Figure 6.

```
Algorithm ApplyfrequentSets_ToTags ()    /* Aspect & opinion Extraction */
// Input:   PSL – Positive Seed List
          NSL – Negative Seed List
          W – Tokenized sentence
          T – Tagged sentence
          i – Current word/tag index
          AI – aspect index modifier
          OI – opinion index modifier
// output: aspect – extracted aspect
                 opinion – extracted opinion
2.  listOfTags1 = { " ", " ", " ", … }  /* tags from predefined frequent sets */
3.  listOfTags2 = { " ", " ", " ", … }
4.  for each tag1 in listOfTags1 do
5.     for each tag2 in listOfTags2 do
6.        if Ti ∈ tag1 AND Ti + 1 ∈ tags2 then
7.           if Wi + OI ∈ PSL OR Wi + OI ∈ NSL
8.              aspect = Wi + AI
9.              opinion = Wi + OI
10.          end if
11.       end if
12.    end for
13. end for
```

Figure 6. Extraction algorithm.

## 5. EXPERIMENT AND EVALUATION

### 5.1. Data set

We conducted the experiment using Hu and Liu's dataset [2] consisting of annotated customer reviews of five different products: (Canon camera, DVD player, MP3 Player, Nikon and Nokia). These reviews, written by different customers , were collected from Amazom.com and Cnet.com and processed by Hu and Liu in [2]. The reviews contained 2,500 sentences.  Each dataset consisted of more than 260 sentences found to be opinionated reviews written by 325 customers.  The format of the datasets is unstructured text files.  To evaluate the discovered aspects, a human tagger manually read all of the reviews and labelled aspects and associated opinions for each sentence. Before , we use the datasets, we pass the dataset to a pre-processing filter to remove all humane annotations and keep the original collected reviews.

### 5.2. Evaluation Criteria

To evaluate the performance of the proposed technique, we adopted three measurements named, precision, recall, and f-measure, and then we compared these measures to the baseline model proposed by Hu and Liu [2].The evaluation involved two perspectives: the effectiveness of aspect extraction and opinions extraction processes.

## 6. RESEARCH RESULTS

Having completed the aspect and opinion extraction, we reviewed our results.  As shown in Table 5, our framework yielded improved precision and maintain the same recall compared with the novel work proposed by HU & Liu in  [2].

Table 5 shows the average precision and recall of the five products reviews named (Canon camera, DVD player, MP3 Player, Nikon and Nokia), along with the calculated f-measure of precision and recall. The precision reflects the ration of accuracy of classified aspects and opinions to the number of all reviews, while recall reflects the ration of completeness of all reviews classified correctly.

Table 5.Comparison of proposed technique and baseline model

| | Average Precision | |
|---|---|---|
| | Aspect extraction | Opinion extraction |
| Baseline [2] | 0.7 | 0.64 |
| Proposed technique | 0.99 | 0.56 |
| | **Average Recall** | |
| | Aspect extraction | Opinion extraction |
| Baseline [2] | 0.79 | 0.69 |
| Proposed technique | 0.64 | 0.61 |
| | **F-measure** | |
| | Aspect extraction | Opinion extraction |
| f-measure for Baseline [2] and Proposed technique | 0.74 | 0.65 |
| | 0.77 | 0.60 |

From previous results, we conducted t-*tests* to quantify the improvement of precision and recall for the extraction processes. The value of the t-*test* for precision for aspect extraction is *"0.0001"* and for recall for aspect extraction is *"0.0172"* which considered being extremely statistically significant. The value of the t-*test* for precision for opinion extraction is *"0.0851"* and for recall for opinion extraction is *"0.0941"* which performs normally compared to the baselines model and leave us with a room to improve the opinion extraction.

## 7. CONCLUSION AND FUTURE WORK

In this paper, we proposed framework to produce an opinionated summary from customer reviews. The main achievement involved the task of aspect and opinion extraction. The extraction was based on data mining, natural language processing and ontology techniques. The main objective of this study is to provide "aspect-based opinionated summary" from customer reviews of online sold products. Our experimental results showed great promise for the technique. At this stage, we achieved very high precision and a normal recall performance compared to the baseline model in extracting aspect and opinion .In future work, we plan to improve and enhance our technique to achieve higher results.

## 8. REFERENC